\documentclass[11pt,a4paper]{article}
\usepackage[hyperref]{eacl2021}
\usepackage{times}
\usepackage{latexsym}
\usepackage{xspace}

\PassOptionsToPackage{hyphens}{url}\usepackage{hyperref}

\usepackage{microtype}

\usepackage{csquotes}
\usepackage{graphicx}
\usepackage{hyperref}
\hypersetup{
    breaklinks=true
}

\aclfinalcopy 

\newcommand{\eg}{e.g.,\xspace}

\newcommand{\ie}{i.e.,\xspace}

\newcommand{\cf}{cf.}

\newcommand{\me}{\textit{M}}
\newcommand{\sd}{\textit{SD}}

\def\multitasker{{DialoGPT (MT)}} 
\def\edneil{{ED$\rightarrow$RDG}}

\title{Estimating Subjective Crowd-Evaluations as an Additional Objective\\ to Improve Natural Language Generation}

\author{Jakob Nyberg \\
  KTH Royal Institute \\ of Technology, Sweden \\
  \texttt{jaknyb@KTH.se} \\\And
  Ramesh Manuvinakurike \\
  Intel Labs, USA \\
  \texttt{ramesh.manuvinakurike@}\\\texttt{intel.com} \\\And
    Maike Paetzel-Prüsmann \\
  University of Potsdam, Germany \\
  \texttt{paetzel-pruesmann@}\\\texttt{uni-potsdam.de} \\}

\date{}

\begin{document}
\maketitle
\begin{abstract}

Human ratings are one of the most prevalent methods to evaluate the performance of natural language processing algorithms. Similarly, it is common to measure the quality of sentences generated by a natural language generation model using human raters.
In this paper, we argue for exploring the use of subjective evaluations within the process of training language generation models in a multi-task learning setting.
As a case study, we use a crowd-authored dialogue corpus to fine-tune six different language generation models. 
Two of these models incorporate multi-task learning and use subjective ratings of lines as part of an explicit learning goal. 
A human evaluation of the generated dialogue lines reveals that utterances generated by the multi-tasking models were subjectively rated as the most typical, most moving the conversation forward, and least offensive. 
Based on these promising first results, we discuss future research directions for incorporating subjective human evaluations into language model training and to hence keep the human user in the loop during the development process.

\end{abstract}

\section{Introduction}

Creating spoken dialogue systems includes a multitude of challenges as they involve various language processing  (NLP) components. One such important component concerns natural language generation (NLG).
Traditionally, the performance of a NLG unit has been evaluated using automatic metrics, such as BLEU \cite{bleu} or METEOR \cite{meteor}. 
Human evaluations of NLG (i.e., rating autonomously generated dialogue responses) is still the most  common (see \newcite{li-etal-2016-deep,empathetic_dialogues, hashimoto2019unifying,dialogpt} for measuring the performance of such approaches).
Comparing automatic metrics with human evaluations, however, has shown little correlation between the two \cite{how-not-to-evaluate, automatic-turing, belz-reiter-2006-comparing, novikova-etal-2017-need, reiter-2018-structured}, which stresses the importance of using human evaluations to rate the suitability of a system or part of a system that will ultimately be used by humans again. 
In recent times, appreciable advances have been made in developing automated metrics showing  correlation with the human ratings \cite{zhang2019bertscore,mehri2020usr}. These approaches, however, do not provide a method for measuring the affect and emotional aspects of the generated content which is central to our approach.

Despite human evaluations becoming increasingly prevalent and affordable, 
they are usually only seen as the final stage of the system design process. Evaluations are hence performed after concluding the implementation work and used to compare the new approach to previous models or techniques. 
The resulting feedback from the users 
is then discarded unless used for future comparisons.
In this paper, we argue for \textit{keeping the human user in the loop by including human evaluations in subsequent natural language generation processes.} 
To keep the development of such a system at a fast pace and low overhead cost, human evaluations can not rely on a few experts but need to utilize online crowd-workers. While crowd-sourcing platforms allow us to gather ratings of several hundred dialogue lines within a few minutes, such evaluations cannot rely on sophisticated metrics requiring a high skill or long training process of the raters but need to utilize subjective ratings of lines instead. 

To the best of our knowledge, this paper is a first proof-of-concept exploration to \textit{include subjective utterance ratings from crowd-workers collected at a low cost and in a short time during the training of a system which is generating responses for a dialogue agent.} As a domain, we use the geography-themed cooperative guessing game RDG-Map in which a human and an embodied conversational agent try to identify countries on a world map \cite{paetzel2019digo}. To enhance the social component of the dialogue, the human-robot team has a brief chat before and after each game. Ultimately, we aim to increase people's engagement playing with the agent by adapting its behavior to the human player. Depending on the learning and playing style of a person, the agent should maximize the team's performance by either encouraging or challenging the team mate during play.
As a first step, the agent was given two affect states based on \newcite{russell1980circumplex} which influence its dialogue behavior: In addition to an indifferent behavior, utterances could be \textit{excited and encouraging} or \textit{impatient and provocative}. Depending on the team's performance in the game and the human responses to the agent, the affect state of the agent gradually changes over time.

Initially, we used a crowd-authoring technique to gather potential responses for our dialogue agent \cite{mota2018expressing}. It has previously been shown that such crowd-authored content helps achieve variety in a dialogue system's responses \cite{wang2012crowdsourcing,mitchell2014crowdsourcing,shah2018bootstrapping}. To design the system described in this paper, we first gathered subjective evaluations by a different set of crowd-workers, rating the dialogue utterances on the dimensions typicality, offensiveness, and affect. We then used these utterances for training models for neural language generation. We trained six model variations to generate responses for different combinations of game scenario descriptions and affective state. Two models were trained using \textit{multi-task learning goals, making the estimation of the subjective affect rating of the utterance a secondary goal of the model.} The main contribution of this paper is the \textit{performance analysis of the multi-task models trained on crowd-sourced evaluations compared to the models solely tasked with generating dialogue lines}. To compare the different models, they were used to generate utterances for scenarios both seen and unseen during training and resulting dialogue lines were then fed back into the evaluation system, acquiring the same human evaluations obtained for the original crowd-authored lines. In addition to analyzing differences in subjective ratings of the dialogue lines, we compare the human evaluations to the BLEU score as an example of a traditional automatic metric for evaluating language generation models. We conclude the paper by discussing advantages and challenges of our human-in-the-loop language generation pipeline and suggest future work to improve upon and further evaluate the suitability of our proposal.

\section{Related Work}

The role of crowd-workers in the development of NLG models can be two-folded: Sentences provided by crowd-authors can be utilized as a source of the surface form of the sentences that the dialogue system needs to generate or as feedback about the performance of the NLG model. Methods for crowd-sourcing content include: \textbf{(i)} requesting the users to generate a sentence given a context \cite{duvsek2016context}, \textbf{(ii)} asking users to generate surface forms using templates \cite{wang2012crowdsourcing,mitchell2014crowdsourcing}, and \textbf{(iii)} showing the dialogue to crowd-workers and asking them to paraphrase a given dialogue \cite{shah2018bootstrapping}. Utterances collected using these approaches have been shown to be diverse and have been used to train neural NLG models, some of which have achieved impressive results in recent times. Another method to utilize crowd-sourcing is to request crowd-workers to rate the generated sentences on various performance metrics \cite{dethlefs2012optimising,rieser2014natural}. Recent works have studied utilizing human evaluations to train neural models directly \cite{gpt-preferences}. Human judgments were shown to be particularly useful for machine learning tasks where the loss function for the intended learning goal is difficult to express with the data alone. The related work, however, did not focus on dialogue generation but on other tasks that are difficult to quantify objectively, like summarization. 

While recent prominent neural NLG models have been able to generate human-like sentences, they are not only very large (in terms of the number of parameters), but also trained on enormous data sets (in terms of the number of training samples)  \cite{transformer, ctrl, gpt, gpt2, gpt-3, comparison}.
Such models can respond well even in challenging dialogue tasks \cite{dialogpt, meena, challenges}. Due to the hardware and data requirements of such models, fine-tuning pre-trained models is a popular approach for obtaining well-performing language generation models \cite{howard2018universal,chen2020distilling,huggingface_transformers,he2021analyzing}. Lack of consistency is one of the major issues in neural dialogue generation, which has been tackled by methods such as including persona or situation description to improve the consistency between generated sentences across multiple turns of dialogue. \cite{persona-chat, liu2020you, transfertransfo}. 
In a similar fashion, the question of how to incorporate information that enables the consistent generation of affective, empathetic, or emotional dialogue has been extensively studied \cite{emptransfo,shen2020cdl,mojitalk,qian2018assigning,lubis2018optimizing, empathetic_dialogues, moel}. 

In this work, we extend the literature by exploring an approach for developing an NLG pipeline using crowd content and subjective evaluations for a limited corpus of in-domain data. Following \newcite{intermediate}, we leverage the \textit{EmpatheticDialogues} (ED) corpus by \newcite{empathetic_dialogues} as an intermediate training step before training on the domain-specific data. 
We apply models by \newcite{gpt2} and \newcite{dialogpt} on the crowd-sourced content and human evaluations to generate utterances for the given domain.
Like in the works of \newcite{transfertransfo}, \newcite{emptransfo} and \newcite{gpt-preferences}, we use pre-trained models to reduce the amount of hardware and crowd-sourced data needed. However, we do not use human judgments for reinforcement learning, like \cite{gpt-preferences} or \cite{nguyen-etal-2017-reinforcement}, but for supervised learning.

\section{A Crowd-Sourced Human Evaluation Pipeline}
\label{sec:hum_eval}

Our pipeline to collect crowd-sourced ratings of dialogue lines follows the approach described by \newcite{mota2018expressing} with few alterations. In the first evaluation stage, a set of untrained crowd-workers are asked to judge how typical and ordinary a sentence is given a situational description and how offensive it is on a five-point Likert scale. They are also asked if the utterance is nonsensical, in which case the relevancy and offensiveness questions are skipped. The second evaluation stage focuses on the affect of utterances, and workers are asked to judge whether a sentence is excited, frustrated or indifferent. In case they perceived the sentence as excited or frustrated, they need to mark the strength of the affect on a scale from 1 (slightly) to 4 (extremely). For easier computation going forward, the affect rating is combined into a single scale ranging from -4 to +4, with negative values indicating frustration, 0 indicating indifference, and positive values indicating excitement.

The pipeline runs fully automatically, given a set of input utterances. Each new task that is created and uploaded to Amazon Mechanical Turk consists of five utterances and is rated by five different crowd-workers. Crowd-workers are allowed to take multiple tasks in a row, which results in a varying level of familiarity with the task of individual raters. Once evaluations for the first and second stage have been performed by five people, their scores are automatically aggregated into a single average rating per line. Figure \ref{fig:example_utterances} shows a sample evaluation of a line written by a human crowdworker and three language generation models for a given scene. 

Crowd-workers were required to be based in the US and have an approval rate of at least 80\% to take our HITs. They received \$ 0.15 USD per task they completed. Participation was fully anonymous and no personal data was collected. People who responded randomly to our task (see Section \ref{sec:discussion} for a discussion) were manually flagged as unreliable. Their ratings were consequently removed from the result aggregation, and a respective number of replacement tasks were uploaded.  
    
\section{Model Implementation and Training}


\subsection{Training Corpora}

Two sets of corpora were used in this project: The set of utterances collected and rated by crowdworkers specifically for the RDG-Map game, and the
\textit{EmpatheticDialogues} (ED) corpus by \cite{empathetic_dialogues}.
 EmpatheticDialogues was used as an intermediary training step, with some models being trained for response generation on ED before being fine-tuned to the RDG-Map data (denoted as \edneil{}) to give the models time to learn the syntax of the task on a large dataset before applying them to the small domain-specific corpus. 

\paragraph*{EmpatheticDialogues Corpus}
\label{sec:ED}
EmpatheticDialogues is a corpus which consists of 24850 conversations that are connected to a textual description of a personal experience \cite{empathetic_dialogues}. Crowdworkers were asked to describe a situation in which they felt one of 32 given emotions. Two crowdworkers then conversed about their experience for up to six dialog turns. Unlike the RDG-Map data, ED is not evaluated by human raters. Instead, the dialogue is assumed to match the designated situation.


\paragraph*{The RDG-Map Corpus and Its Crowd-Sourced Affective Evaluations}
\label{sec:neil_data}
The RDG-Map data was collected using the crowd-sourcing process described previously. The aim of the dataset is to expand the original behavior of the dialogue agent to make the interactive experience more engaging. The dataset consists of 1512 utterances associated with 61 different scenarios that occur in the RDG-Map game and the pre- and post-game social chat. Each scenario has a description of the situation the human and robot are currently in and a direction for the next utterance to be authored for the robot (\cf~\autoref{fig:example_utterances} for a sample). 
Each scenario includes three different target \textit{affects}: The robot is described as either \textit{excited and encouraging}, \textit{impatient and provocative}, or \textit{indifferent}.

 The RDG-Map corpus resembles ED in its main characteristics: ED includes situational descriptions, emotional labels, at least one dialogue line per scenario, and comparable data fields. However, several notable differences exist between the two corpora: For ED, the emotion label refers to an experience rather than the content of the dialogue line, and the description of the experience is narrated in first-person instead of the third-person format of the RDG-Map scenarios. Moreover, the situational descriptions in ED refer to a prior event rather than a current situation. Perhaps the most notable difference that for ED, the affect is recorded as textual emotion labels, whereas for RDG-Map, it is recorded as a value. This means that in order to perform emotion prediction on both sets, either the task has to be changed between the two sets, or the data has to be converted. This is explained further in Section \ref{sec:multitask}.

\begin{figure}[!bt]
\centering
\fbox{\begin{minipage}{\linewidth}
\textbf{Scenario:} The human and the robot have finished playing the game
and talked about the game for a little while. 
If the robot is \textit{excited}, how would it say goodbye to the human player? \\

\textbf{Human:} I've got to go. Goodbye. \\ (Typicality: 3.4, Offensiveness: 1.6, Affect: 0.0) \\
\textbf{RDG:} Good to meet you, human. See you around. \\ (Typ: 4.2, Off: 1.6, For: 3.8, Aff: -1.0) \\
\textbf{ED$\rightarrow$RDG:} You did so well, you did so so well! \\ (Typ: 4.2,	Off: 2.2, For: 4.4, Aff: 3.4) \\

\end{minipage}}
\caption{Responses to a sample scenario, produced by a human crowdworker and \multitasker{} trained with different sets of data, with human evaluation scores shown underneath. Explanations of scores can be found in Sections \ref{sec:hum_eval} and \ref{sec:analysis}.}
\label{fig:example_utterances}
\end{figure}

\subsection{Language Generation Models}
\label{sec:model_variaions}

Three variations of pre-trained transformer-based response generators were trained with the collected utterances: GPT-2 \cite{gpt2}, DialoGPT \cite{dialogpt} and DialoGPT with multitasking (further on referred to as \enquote{\multitasker{}}\footnote{MT in this scenario refers to \enquote{Multitasking}, and not \enquote{Machine Translation} which is also commonly abbreviated as \enquote{MT}}) . 
These three models were in turn trained with two levels of fine-tuning, either being trained only on RDG-Map data or first on EmpatheticDialogues followed by RDG-Map data. This led to a total of six model variations.
Worth noting is that GPT-2 and DialoGPT are architecturally the same model, both being decoder-only transformers but trained on different sets of data. The only architecturally different variant is \multitasker{}, which adds two parallel output layers.

All training was done using the ParlAI framework \cite{parlai}. Implementations, configurations, and pre-trained parameters for GPT-2 and DialoGPT were sourced from HuggingFace's Transformer library \cite{huggingface_transformers}. All models are \enquote{medium} sized models with 24 attention layers, which amounts to about 345 million trainable parameters and a vocabulary of 50000 tokens. 

\subsection{Decoder Selection}
We considered three decoding methods for our language model: greedy decoding, top-$k$ sampling and nucleus sampling \cite{degeneration}. 
\multitasker{}, trained with \edneil{}, was used to generate utterances with the three decoding methods, since it had the lowest perplexity on the evaluation data set. Scenario and affect combinations were selected in the same way as described in Section \ref{sec:analysis}. Five sentences per scenario and affect were generated for top-$k$ and nucleus sampling (total: 90) and one utterance per context was evaluated for the greedy decoding (total: 30) since it always generates the same utterance for a given context. 

Evaluation of utterances were done using the questions described in Section \ref{sec:hum_eval}, measuring typicality, offensiveness and affect. A statistical analysis of the ratings found that top-$k$ decoding produced the most typical and least offensive output, by a slight margin compared to greedy decoding. Affect ratings did not differ significantly between the decoding methods. However, top-$k$ produced the widest range of affect, which led us to use it for the main evaluation.



\subsection{Learning Goals}



\label{sec:multitask}
For GPT-2 and DialoGPT without multi-task training, the only training goal was to predict the human-written utterance associated with the given context, i.e., the game situation with the affective state. \multitasker{} also does this, in addition to two further training goals that contribute to the total training loss.
To include the affect score from the human evaluations during training, an emotion classification task was included following the example of \newcite{emptransfo}. The classification head consists of a single linear layer with dropout. The task varied slightly between the two data sets. When training on RDG-Map data, the head estimated the average affective evaluation score of the utterance, which represents how excited or frustrated it was perceived as. The evaluation score is a decimal value in the range [-4, 4]. When training on EmpatheticDialogues, the head classified the input into one of 32 emotion categories. Because of the different number and types of emotion labels between EmpatheticDialogues and the RDG-Map data, the prediction head could not be preserved from one training phase to the next. The layer was thus re-initialized when switching data sets. A potential solution to this issue, not implemented in this work, would be to predict embedding vectors representing the emotion content in the input, similar to those in \newcite{deepmoji}.

Following the works of \newcite{transfertransfo} and \newcite{bert}, \textit{next-sentence prediction}, or \textit{multiple choice}, was also used as another learning objective for \multitasker{}. The idea of next-sentence prediction is to train NLP models to associate connected parts of the input, such as one turn of dialogue preceding another, to improve the coherence of the generated text.
In our implementation, the task worked as follows: Along with the true utterance written by a human for a specific scenario, a random utterance from another scenario was picked. The model was then presented with both utterances and tasked with deciding which one is the actual response to the scenario. 

\section{Analysis}
\label{sec:analysis}
The performance analysis of the two models utilizing multi-task learning in comparison to the four models trained with the sole task of generating dialogue lines was based both on automated metrics as well as a second round of human evaluations. 

To get a first estimate of how well the models predict the RDG-Map data, the average per-token perplexities of the models on the test set were recorded. We also calculated the average BLEU score for utterances generated from scenarios in the test set. For each generated utterance, all corresponding lines written by humans for that specific combination of scenario and affect were used as references.

For the human evaluation of the different models, a set of utterances to be evaluated was generated. All models used top-$k$ decoding with $k=25$. Six scenarios (three seen and three unseen during training) were used for testing, with three affect categories each (excited, indifferent, and impatient). Each model generated five utterances for each of the six scenarios with the three affects. Each model thus had 90 utterances evaluated, for a total of 540 utterances across all models.

The evaluation pipeline described in Section \ref{sec:hum_eval} was used to gather human ratings of the utterances generated by the language models. One additional question was added to the first stage, asking crowdworkers to rate how much the given utterance moves the conversation forward.  
258 workers participated in the evaluation. Each worker participated in 4 different tasks on average, with a standard deviation of 10 tasks.
\section{Results}


\subsection{Performance of Multiple Training Goals}

Since the multitasking model implemented two additional classifiers, the accuracy of these were tested. For the multiple-choice task, the model trained with ED$\rightarrow$RDG picked the correct label with an accuracy of 82\%, whereas the model only trained on RDG-Map data had an accuracy of 55\%.

To calculate the accuracy of the emotion estimation head, the output was rounded to the closest integer between -4 and 4. This makes the output match the evaluation form shown to crowd workers, where utterances are classified as either excited, neutral or frustrated. The F1 scores of ED$\rightarrow$RDG model were higher than those of RDG.
For both models, the F1 scores for classifying neutral utterances were lower than for the other labels. This is to be expected given the proportions of the training data, as utterances evaluated as neutral are rare, and those rated as excited are the most frequent.

\begin{table}[]
\caption{F1 scores on test set (242 utterances) for multitasking models.}
\begin{tabular}{llll}
\textbf{Data}      & \textbf{Excited}    & \textbf{Neutral}    & \textbf{Frustrated} \\
\hline
ED$\rightarrow$RDG & 0.96 & 0.29 & 0.99  \\
RDG    & 0.93 & 0.00         & 0.96
\end{tabular}
\end{table}


\begin{table*}[]
    \centering
    \begin{tabular}{lllrrrr}
    \textbf{Model} & \textbf{Data} & \textbf{Rating}       & \textbf{Max.} & \textbf{Min.} & \textbf{Mean}  & \textbf{Std. Dev.}  \\
    \hline
    \multitasker{} & ED$\rightarrow$RDG &  Excited	    & 3.6  & 0.2  & 1.6 & 1.0  \\
\multitasker{} & ED$\rightarrow$RDG &  Frustrated & 3.8 & 0.2 & 1.5 & 1.1 \\ 
    \multitasker{} & RDG    &             Excited    & 3.6  & 0.2  & 1.2 & 0.9 \\
\multitasker{} & RDG                &   Frustrated & 3   & 0.2 & 1.0 & 0.7 \\
 Human      & 	& Excited& 3.8 & 0.2 & 1.5 & 0.9  \\
  Human      & &	Frustrated& 4 & 0.2 & 1.4& 0.9 \\
    \end{tabular}
    \caption{Affective ratings for utterances produced by multitasking model. Human ratings for comparison. Scores range from 0 to 4, with 0 indicating indifference.}
    \label{tab:stage3}
\end{table*}




\subsection{Evaluation of the Model Performance}
A two-way ANOVA with the \textit{model} (DialoGPT, \multitasker{} and GPT-2) and the \textit{training set} (ED $\rightarrow$ RDG, RDG) as independent variable was performed using both the BLEU score and the human evaluation as dependent variables.

\paragraph*{Automated Metrics}
The data did not show a significant influence of the model, $F(2, 501) = 0.42$, $p = .658$, or the training data set, $F(1, 501) = 0.16$, $p = .692$, or an interaction effect between the two, $F(2, 501) = 0.82$, $p = .441$, on the generated lines. The BLEU score of the utterances is, however, significantly positively correlated with the crowdworker rating of typicality, $\rho = 0.137$, $p = .002$, and how much the lines advances the conversation, $\rho = 0.106$, $p = .017$.

\paragraph*{Human Evaluation}
Ratings from crowd-workers showed that both the model, $F(2, 534) = 32.13$, $p < .001$, and the training data, $F(1, 534) = 100.41$, $p < .001$, significantly influenced how typical and ordinary the generated lines were perceived. Using a Tukey's PostHoc test, we found that the \multitasker{} model was rated as the most typical ($\me = 3.27$, $\sd = 0.05$) compared to both DialoGPT ($\me = 2.76$, $\sd = 0.05$), $p < .001$, and GPT-2 ($\me = 2.87$, $\sd = 0.06$), $p < .001$. The difference between DialoGPT and GPT-2 was not significant, $p = .218$. There was also a significant interaction effect between the model and the data set it was trained on, $F(2, 534) = 16.35$, $p < .001$. A PostHoc test suggests the existence of two groups of models that perform almost identical: If any of the models was only trained on RDG-Map data, the performance between models was comparable. When including the EmpatheticDialogues data, only \multitasker{} reached the same level of performance. DialoGPT and GPT-2 trained on ED$\rightarrow$RDG both fell in the low-performing group compared to the other combinations.

A similar result was obtained for the crowdworker rating of how much each line moves the conversation forward. Again, both the model, $F(2, 534) = 9.789$, $p < .001$, and the training data set, $F(1, 534) = 112.515$, $p < .001$, had a significant influence on the ratings. \multitasker{} was found to be the model that generated the lines advancing the conversation most ($\me = 3.54$, $\sd = 0.04$) and the difference was significant in comparison to both DialoGPT ($\me =  3.33$, $\sd = 0.04$), $p < .001$, and GPT-2 ($\me = 3.35$, $\sd = 0.05$), $p < .001$. The difference between DialoGPT and GPT-2 was not significant, $p = .925$. Using only the RDG-Map data set for training ($\me = 3.64$, $\sd = 0.03$) generated lines that were perceived as advancing the conversation more than when the models were trained on the EmpatheticDialogues data in addition ($\me = 3.18$, $\sd = 0.03$). An interaction effect between the model and the training data could be observed as well, $F(2, 534) = 33.022$, $p < .001$, which showed a significance between the same two groups of well performing (all models trained on the RDG-Map data set plus \multitasker{} trained on ED$\rightarrow$RDG) and low performing variations (DialoGPT and GPT-2 trained on ED$\rightarrow$RDG).

The model, $F(2, 534) = 12.46$, $p < .001$, but not the data set it was trained on, $F(1, 534) = 1.03$, $p = .31$, significantly influenced the rating of offensiveness of the utterances that were generated. \multitasker{} generated the least offensive lines ($\me = 2.43$, $\sd = 0.05$) in comparison to DialoGPT ($\me = 2.66$, $\sd = 0.04$), $p < .001$, and GPT-2 ($\me = 2.72$, $\sd = 0.05$), $p < .001$. The ratings between DialoGPT and GPT-2 were comparable, $p = .639$. The interaction effect between the model and the data it was trained on was significant again, $F(2, 534) = 16.01$, $p < .001$. This time, the best performing models were the \multitasker{} trained on both RDG-Map alone and the ED$\rightarrow$RDG combination, as well as DialoGPT trained on ED$\rightarrow$RDG.  

Both the model, $F(2, 534) = 12.548$, $p < .001$, and the data set, $F(1, 534) = 2.189$, $p = 0.14$, had a significant influence on the affective ratings of the lines. \multitasker{} produced lines that were on average rated as more excited and encouraging, which is significant compared to lines generated by DialoGPT, $p < .001$, and GPT-2, $p < .001$. The \multitasker{} was also the model that generated lines that covered the most diverse affect in comparison to the other two. The models trained on the ED$\rightarrow$RDG combination were more frustrated and provocative compared to the models trained on the RDG-Map data alone. The combination of model and data set was significant as well, $F(2, 534) = 13.224$, $p < .001$. The three models rated on the more excited end of the affective scale were the two \multitasker{} models and the GPT-2 model trained on the RDG-Map data alone. The most impatient lines were generated by GPT-2 trained on ED$\rightarrow$RDG. A selection of affective ratings is shown in \autoref{tab:stage3}.


\begin{table*}[!htbp]
    \centering
    \begin{tabular}{ll | rr | rrrrrrr}
    & & \multicolumn{2}{c |}{\textbf{Automatic}} & \multicolumn{6}{c}{\textbf{Human Subjective Evaluation}} \\
\textbf{Model} & \textbf{Data}  & \multicolumn{2}{c |}{\textbf{BLEU}} & \multicolumn{2}{c}{\textbf{Forwardness}} & \multicolumn{2}{c}{\textbf{Offensive}} & \multicolumn{2}{c}{\textbf{Typical}}   \\
& & \me & \sd  & \me & \sd & \me & \sd & \me & \sd \\
\hline
\multitasker{} & ED$\rightarrow$RDG   & $0.41$ & $0.3$ & $3.6$ & $0.5$ & $2.5$ & $0.6$ & $3.2$ & $0.6$ \\
\multitasker{} & RDG                 & $0.37$ & $0.3$ & $3.5$ & $0.5$ & $2.4$ &  $0.6$ & $3.3$ & $0.6$ \\ 
Human         &                      & - & - & - & - &  $1.9$ & $0.7$ & $3.4$ & $0.7$ \\
    \end{tabular}
    \caption{Average BLEU scores and ratings for forwardness (\ie{}~moving the conversation forward), offensiveness and typicality for multitasking model. Human ratings for comparison. Typicality ranges from 0 to 5, with 0 representing nonsensical content. Offensiveness and Forwardness range from 1 to 5.}
    \label{tab:stage2}
\end{table*}

\paragraph*{Comparing Language Models and Crowd-Authors}
Eventually, we want to be able to use the language models presented in this paper to generate utterances that are comparable in their rating to the lines authored by crowd-workers. To understand whether our models achieve human-level performance, we combined the model and training set into a single independent variable and tested it against the ratings given to the crowd-authored lines. A one-way ANOVA with a Tukey's PostHoc analysis indeed showed that the ratings of the lines generated by all four models in the high performing group showed no significant difference to the ratings of the human lines, $p \ge .948$ for all four models. The two models in the low-performing group, however, were rated as significantly less typical than the lines written by crowd-authors, $p < .001$ for both models. The affective rating and range of affect between five out of the six combinations and the human model were comparable, $p > .147$ for all models except for GPT-2 trained on ED$\rightarrow$RDG. This specific model and training data combination produced lines that were on average much more frustrated and provocative than the lines written by crowd-authors, $p < .001$. While the typicality of the lines and their affective range was comparable, utterances generated by all six combinations of model and training data were rated as significantly more offensive than the crowd-authored lines, $p < .001$ for all six models. A comparison between DialoGPT (MT) and the crowd-authored lines is summarized in Table \ref{tab:stage2}. All generated utterances and respective evaluation scores are available publicly on GitHub\footnote{\href{https://git.io/JYzq8}{https://git.io/JYzq8}}.

\section{Discussion \& Future Work} \label{sec:discussion}

We trained six variations of neural language generators on crowd-sourced content and evaluations. Our results suggest that \multitasker{}, the model additionally tasked with predicting the subjective evaluations by crowd-workers, produced utterances that were perceived as the most typical, least offensive, and most capable of moving the conversation forward. It also generated dialogue lines covering the widest range of affects, which meets an important goal for the spoken dialogue system of the RDG-Map domain.
Utterances generated by \multitasker{} \textit{reach scores comparable to those given to human-authored lines in the dimensions relevance and affect for scenarios both seen and unseen during training}; in real-time and at a lower cost than the crowd-sourced approach. Based on these results, we consider the multitask learning approach a success.

\paragraph{Utilization of Subjective Ratings}
While our results are promising when it comes to the success of using subjective ratings as a secondary goal in multi-task learning to generate affective dialogue lines, further research is necessary to understand the exact influence of this particular training objective. 
In this work, we added two additional training goals in order to further utilize the collected data: Multiple choice and emotion classification. 
Hence, it may be possible that the multiple-choice task was more influential for the success of the \multitasker{} model. 
However, in observing the training process, it was noted that the training loss for the multiple choice task decreased significantly faster than the loss of the emotion prediction task. This indicates both that the emotion prediction tasks is a more difficult task to train, and that it plays a larger role during the optimization as its loss term is present during a larger portion of the training process. While future work is necessary to determine the contribution of each task individually, our results show \textit{a strong indication that the inclusion of the subjective ratings contributed more to the performance improvements than distinguishing between a real or fake response}. 

\paragraph{Keeping the Human Rater in the Loop} In this proof-of-concept, we only utilized the initial crowd-evaluations of dialogue lines authored by other humans for training our NLG models. An interesting topic for future exploration would be to further include the second round of evaluations collected for the sentences generated by the NLG models. We could then envision natural language generation as an iterative process, defined by a number of alternating training and evaluation sessions, where models can be adjusted based on the evaluations. This moves the process closer to \textit{reinforcement learning}, which is a topic that has been covered in previous work \cite{li-etal-2016-deep, gpt-preferences, nguyen-etal-2017-reinforcement}. One of the challenges with this approach is finding a reward function which correlates the human evaluations with the content and prevents the model from veering off topic, but with the benefit that the model can be trained on only evaluation data going forward.

\paragraph{Addition of Further Tasks during Training} 
Given the performance improvements offered by multitask learning, a potential subject of future work is to expand the multitasking further and incorporate more of the available human evaluation data. The offensiveness or typicality score are present in the data but are currently unused during training. Utterances rated too low in typicality or too high in offensiveness in the original spoken dialogue system were not included in the agent's conversational corpus. We chose to include rejected lines in the model training data to preserve as much of the problem-specific data as possible. Even if an utterance has been rejected as offensive, it may still relate to the context, which is information that the model theoretically can utilize. However, we found all our models to generate lines significantly more offensive than the original crowd-authored lines. While this finding is in line with related work on DialoGPT, which notes that models trained on large-scale internet text corpora can have issues with producing offensive content \cite{dialogpt, comparison, parrots}, we would still like to limit offensive content in a dialogue system deployed to converse with people. A potential improvement to the training procedure would be to remove rejected lines from training data. 
Another approach would entail the inclusion of typicality or offensiveness in the input which could potentially improve performance. Including the scores might also enable a method of controlling the typicality or offensiveness of the output, like the affect might currently do. It would also be prudent to study to what extent the designated affect actually influences the actual output.

\paragraph{Correlation between Human Evaluations and BLEU}
Contrary to findings in the related work, we found the BLEU score of the individual utterances to be significantly correlated with the human evaluations on typicality and how much the utterances advance the conversation. \newcite{how-not-to-evaluate} note that for constrained domains, the BLEU score correlates better with human judgements, which the RDG-Map domain might be considered as.  However, no correlation could be found between the subjective rating of offensiveness and the automatic metric. This makes sense considering that BLEU is a measure of content similarity, and minor changes to the content, like an exclamation mark, may cause major changes in the perceived offensiveness of an utterance.

\paragraph{Filtering of Evaluations} One major issue we experienced in our crowd-evaluation pipeline concerns the dishonesty of a few crowd-authors who did not pay attention to the task they accepted. While most participants performed their tasks well, a few workers providing nonsensical or random ratings can tilt the results, especially if these workers participate multiple times. To account for this, filters flagging random answers are necessary. This is complicated by the fact that the questions asked in the form are subjective, e.g., how offensive or typical an utterance is perceived. It is thus difficult to verify if people responding randomly as there are no \enquote{correct} answers. A method to address the issue at least partially is to include an objective control question. However, there are challenges around the number of such control questions to include \cite{liu2013scoring} and efficiency of such trap methods \cite{lloret2013analyzing,kittur2013future} for complex NLP tasks. 

Our method to detect crowd-raters responding randomly was to manually examine the results and exclude workers that gave obviously repetitive answers, e.g. always answering with the same alternative throughout multiple tasks. This is a simple but flawed method as raters answering in a random or less predicable, but still disingenuous, manner are not caught through this method. Additionally, our method only works with crowd-workers participating in several tasks. A measure that is simple to enforce is to prevent workers from participating more than once and hence limit the individual influence of each worker. However, this may lead to workers avoiding the task since it is only profitable for them to engage in repeated tasks, and also the loss of workers that give honest evaluations for multiple sessions.
A more refined and automated method of filtering answers would improve the validity of the evaluation scores, and thus by proxy improve the training procedure.

\subsection{Ethical Issues Regarding Language Models}

There are several ethical issue with large-scale language models worth discussing. We observe some of the issues brought up by \newcite{parrots}, the main one being that the output can easily be misinterpreted as coherent or intelligent. One should be careful not to over-attribute the writing capabilities of language models as being equivalent to that of a human, despite in this case being rated similarly to human writers. In this scenario, we tell the raters that a robot produced the utterances, which likely influenced their judgment of typicality. A line assumed to be written by a machine might be considered typical even if it is vague or contains unusual wording, since the rater may consider the language capabilities of a machine to be limited. For future studies into dialogue generation models, it might be prudent to include harsher judgements of quality than used in the present work, \eg{}~asking the raters to judge the sentence as if it was written by a human, or whether it makes sense logically. 





Another issue brought up by \newcite{parrots} is the possibility of models producing offensive language. While we did notice that the lines generated by the language models were evaluated as more offensive than the crowd-authored lines, a manual review of the dialogue output of the language models did not disclose any slurs or explicitly derogatory statements. The utterance considered the most offensive was \enquote{I hope this game will be your last because it will be your last game for a very long time} which may be interpreted as a threat to the recipient's life. Other utterances considered offensive typically involve accusations of laziness or the human not being very good at the game, which are meaningful given the domain and the affect description of the agent.

\section{Conclusion}


The usage of human-annotated data for training machine learning models is an established practice. In this work, we propose and evaluate the utilization of subjective human evaluations for model training that would otherwise be used merely for evaluation. Our results suggest that by using not only crowd-sourced \textit{content}, but also crowd-sourced \textit{evaluations}, we can increase the performance of our models. We hence argue that future work should explore the inclusion of further subjective ratings and the possibility to make model generation and evaluation an iterative process and hence keep the human in the loop during the development process.

\appendix

\newpage

\bibliographystyle{acl_natbib}
\bibliography{refs}

\end{document}